\documentclass[letterpaper, 10 pt, conference]{ieeeconf}  

\IEEEoverridecommandlockouts                              

\usepackage{booktabs}
\usepackage{indentfirst}
\usepackage[table]{xcolor} 
\usepackage{cite}
\usepackage{multirow}
\title{\LARGE \bf
FAR-Net: Multi-Stage Fusion Network with Enhanced Semantic Alignment and Adaptive Reconciliation for Composed Image Retrieval}

\author{Jeong-Woo Park$^{1}$, Young-Eun Kim$^{1}$, and Seong-Whan Lee$^{1}$
\thanks{*This research was supported by the Institute of Information \& Communications Technology Planning \& Evaluation (IITP) grant, funded by the Korea government (MSIT) (No. RS-2019-II190079, Artificial Intelligence Graduate School Program (Korea University)), and No. IITP-2025 RS-2024-00436857 (Information Technology Research Center (ITRC)).}
\thanks{$^{1}$J.-W. Park, Y.-E. Kim, and S.-W. Lee are with the Department of Artificial Intelligence, Korea University, Anam-dong, Seongbuk-ku, Seoul 02841, Korea.
        {\tt\small \{jeongwoo\_park, 
        ye\_kim, 
        sw.lee\}@korea.ac.kr}}%
}

\usepackage{amsmath} 
\usepackage{graphicx}  

\renewcommand{\thetable}{\Roman{table}}

\usepackage{caption}
\begin{document}

\maketitle
\thispagestyle{empty}
\pagestyle{empty}

\begin{abstract}

Composed image retrieval (CIR) is a vision-language task that retrieves a target image using a reference image and modification text, enabling intuitive specification of desired changes. While effectively fusing visual and textual modalities is crucial, existing methods typically adopt either early or late fusion. Early fusion tends to excessively focus on explicitly mentioned textual details and neglect visual context, whereas late fusion struggles to capture fine-grained semantic alignments between image regions and textual tokens. To address these issues, we propose FAR-Net, a multi-stage fusion framework designed with enhanced semantic alignment and adaptive reconciliation, integrating two complementary modules. The enhanced semantic alignment module (ESAM) employs late fusion with cross-attention to capture fine-grained semantic relationships, while the adaptive reconciliation module (ARM) applies early fusion with uncertainty embeddings to enhance robustness and adaptability. Experiments on CIRR and FashionIQ show consistent performance gains, improving Recall@1 by up to 2.4\% and Recall@50 by 1.04\% over existing state-of-the-art methods, empirically demonstrating that FAR-Net provides a robust and scalable solution to CIR tasks.

\end{abstract}


\section{INTRODUCTION}
Composed image retrieval (CIR) \cite{vo2019composing} retrieves a target image conditioned on a reference image and a textual description of the intended modification. This approach moves beyond traditional image retrieval \cite{chua1994content,qu2020context} in which visual similarity alone is used. CIR instead allows users to specify attributes that should be changed or preserved (\textit{e.g.}, the same outfit but a different color), and this retrieval paradigm has been shown to be useful in e-commerce, visual search engines, and content-based recommendation systems \cite{zhang2020empowering}.

\begin{figure}[htbp]
    \centering
    \includegraphics[width=0.9\columnwidth]{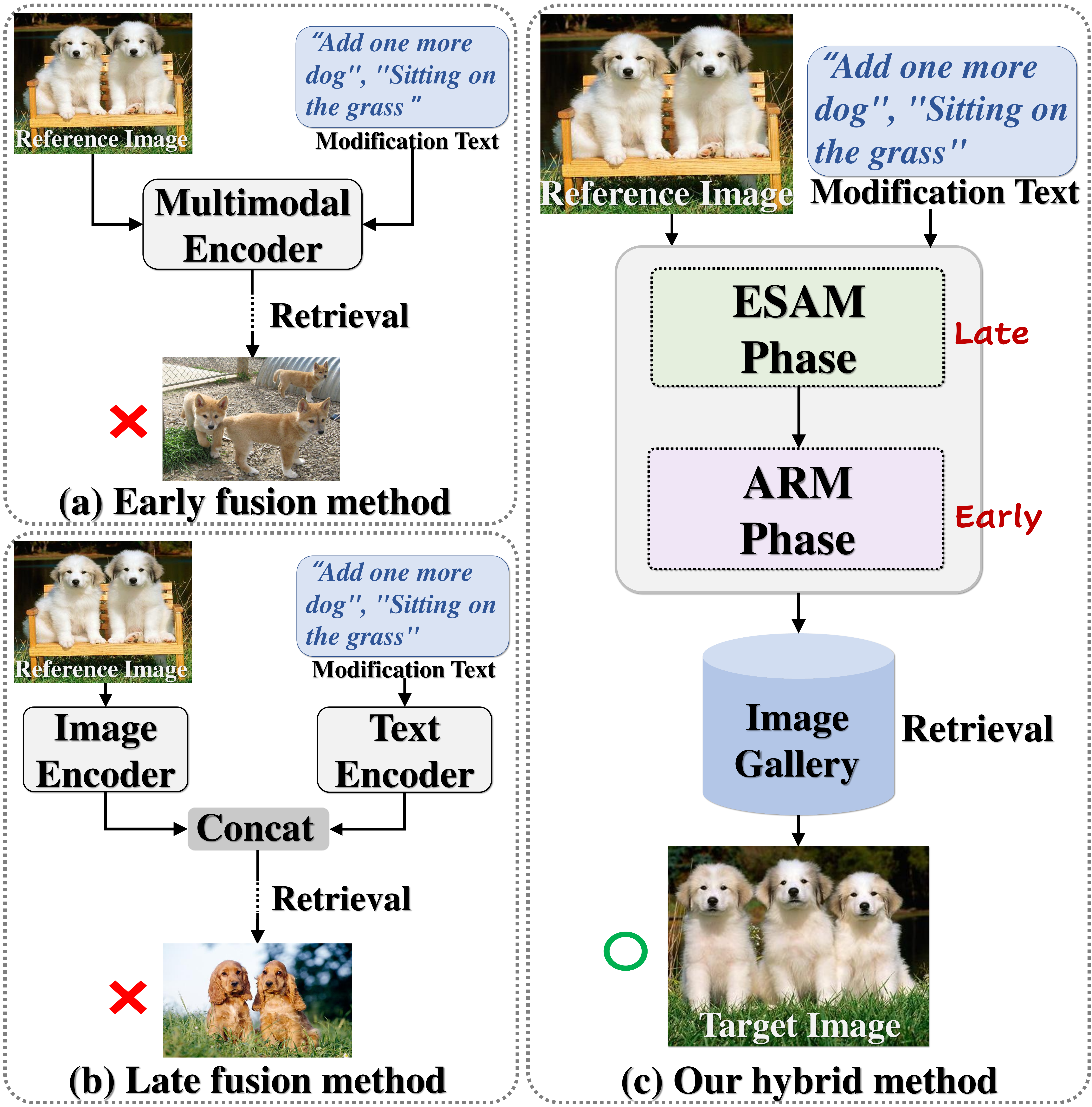}
    \caption{Comparison of existing fusion strategies for CIR. 
(a) Early fusion integrates images and text at a multimodal encoder. 
(b) Late fusion encodes images and text separately before concatenation. 
(c) Our method utilizes late fusion with cross-attention in the ESAM Phase and early fusion with uncertainty embedder in the ARM Phase.
}
    \label{fig:overview_}
\end{figure}

Despite its strong potential for user-intent-driven multimodal search, CIR still poses significant challenges. Unlike image-to-image or text-to-image retrieval, CIR requires integration strategies that jointly process textual and visual inputs with respect to the specific requirements of CIR. To address this requirement, many recent approaches \cite{levy2024data, gu2024compodiff, wen2021comprehensive, wen2024simple, fujisawa1999information} adopt one of two fusion paradigms, as shown in Fig.~\ref{fig:overview_}. The early fusion method \cite{xu2024sentence} processes text and images through a single multimodal encoder, where a cross-attention mechanism enables token-level interactions across both modalities. This method facilitates the precise identification of which parts of an image need to be changed and how. In contrast, the late fusion method \cite{huang2024dynamic} encodes each modality independently and then combines the resulting features, preserving modality-specific characteristics while capturing information at a coarser level.

While each paradigm offers distinct advantages, both present inherent limitations. Early fusion facilitates token-level interactions, enabling the identification of visual elements corresponding to the intended modification text. However, it tends to focus more heavily on components explicitly mentioned in the text \cite{khan2020mmft}, thereby overlooking visually relevant cues that are not explicitly described but may still contribute to accurate retrieval. In contrast, late fusion maintains modality-specific representations and captures global contextual relationships between image and text. Nevertheless, it is suboptimal in scenarios that require precise alignment between textual cues and their corresponding visual components. This limitation arises from the insufficient level of interaction between image and text, which hinders the effective modeling of subtle modifications such as object removals or attribute changes \cite{xu2024sentence,lerner2023multimodal, lee2015motion, lee1997new}.

To mitigate these limitations, we propose FAR-Net, a multi-stage fusion framework that incorporates enhanced semantic alignment and adaptive reconciliation. FAR-Net adopts a principled Late-to-Early cascaded architecture, seamlessly integrating the strengths of both late and early fusion into a unified pipeline.

The first stage, the enhanced semantic alignment module (ESAM), addresses the limited cross-modal interaction in late fusion. While late fusion retains modality-specific features, its separate processing of visual and textual inputs often fails to capture subtle changes described in the modification text. To address this, ESAM uses a Q-Former \cite{li2023blip} to apply cross-attention between the reference image and the modification text, producing attention maps that reflect their semantic match. It then minimizes the discrepancy between the attention maps generated from the reference–modification and target–modification pairs using an attention alignment objective. This objective is used together with a late fusion loss, enabling ESAM to enhance semantic alignment while maintaining modality-specific details.

Building on this, adaptive reconciliation module (ARM) complements early fusion by mitigating its tendency to focus excessively on explicitly mentioned elements in the modification text, potentially overlooking valuable visual cues. To address this issue, ARM adopts an uncertainty‑aware learning scheme that jointly optimizes two objectives. It aligns the fused representation produced by ESAM with the target image embedding while simultaneously injecting uncertainty-scaled stochastic perturbations into the target embedding and applying the same contrastive loss. Optimizing these objectives  discourages brittle reliance on explicit cues and yields more robust vision‑language representations.

To evaluate the effectiveness of the proposed method, we conduct experiments on FashionIQ \cite{wu2021fashion} and CIRR \cite{liu2021image} benchmarks. The results demonstrate that FAR-Net achieves superior retrieval performance and improved robustness compared to state-of-the-art methods.

The key contributions of this study are as follows:
\begin{itemize}
\item We propose FAR-Net, a multi-stage retrieval framework that structurally integrates both late and early fusion through modular ESAM and ARM, enabling an end-to-end multimodal interaction across stages.
\item To address the limitations of each fusion strategy, ESAM leverages cross-attention to enhance semantic alignment, while ARM employs uncertainty modeling to improve robustness under varying conditions.
\item Extensive experiments on CIR benchmarks demonstrate that FAR-Net outperforms state-of-the-art methods, with gains of up to 2.4\% in Recall@1 and 1.0\% in Recall@50.
\end{itemize}

\begin{figure*}[htbp]
    \centering
    \includegraphics[width=\textwidth]{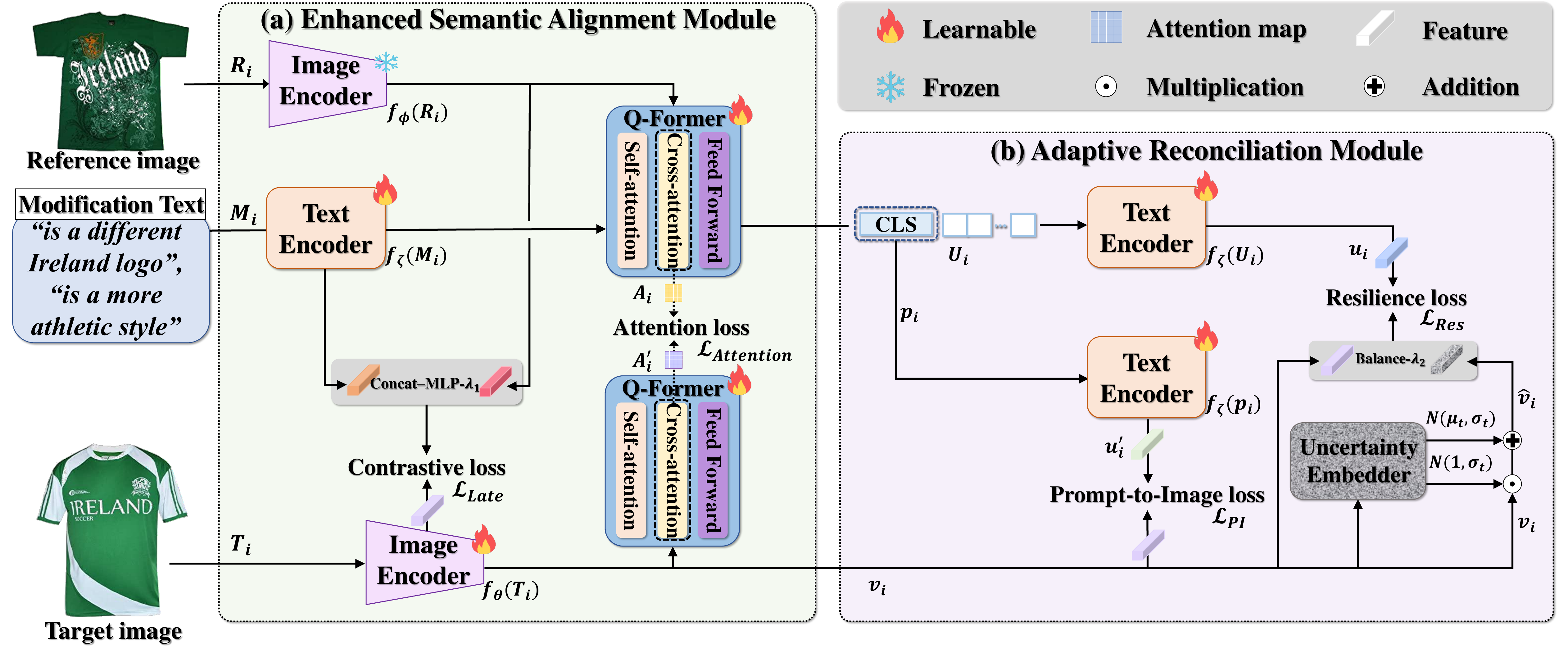} 
    \caption{Overall training pipeline of the FAR-Net framework. 
    This consists of two main modules: (a) The ESAM extracts features from the reference image and the modification text. Subsequently, the concatenated image-text features are optimized using \( \mathcal{L}_{\mathit{Late}} \), and the Q-Former leverages cross-attention mechanisms to refine token-level and region-level semantic alignments guided by \( \mathcal{L}_{\mathit{Attention}} \).
 (b) The ARM refines the fused representation using uncertainty embeddings to account for variability and ambiguity in the data. Perturbed target image representations are generated and balanced with \( \mathcal{L}_{\mathit{Res}} \), while \( \mathcal{L}_{\mathit{PI}} \) ensures alignment between the text prompts and the target image representation.}
    \label{fig:overview}
\end{figure*}

\section{RELATED WORKS}

\subsection{Composed Image Retrieval}
CIR integrates a reference image with textual modifications to retrieve a target image by leveraging pretrained vision-language models that encode and align textual modifications with corresponding visual features~\cite{radford2021learning}. Early CIR approaches \cite{vo2019composing, goenka2022fashionvlp} that leverage contrastive learning often struggled with fine-grained semantics, prompting subsequent methods like CosMo \cite{lee2021cosmo} and CLVC-Net \cite{wen2021comprehensive} to refine feature integration for better compositionality. Advancements in Vision-Language Pre-training (VLP), such as efficient alignment mechanisms like Q-Former \cite{li2023blip}, serve as effective tools for improving CIR performance. Nevertheless, effectively capturing detailed multimodal interactions remains a key challenge in CIR.

\subsection{Fusion Strategies for Composed Image Retrieval}
Fusion approaches are critical in CIR, with research typically adopting either early or late fusion, each presenting distinct trade-offs. Early fusion approaches \cite{xu2024sentence, fu2025pair} facilitate rich cross-modal interactions through joint encoding, enabling fine-grained alignment. However, they risk focusing excessively on textual cues while neglecting crucial visual context. In contrast, late fusion approaches \cite{gu2024compodiff, wen2023target, song2024syncmask} preserve modality-specific characteristics via separate encoding and effectively retain unique unimodal information. However, they often lack the precise semantic alignment needed to model subtle textual modifications. Despite these complementary strengths and weaknesses, most existing CIR methods still rely on a single fusion paradigm, failing to combine their advantages. To address the respective limitations of early and late fusion, FAR-Net introduces a principled Late-to-Early cascaded architecture that structurally integrates complementary strengths, setting it apart from prior works.

\section{METHOD}

\subsection{Problem Formulation}
\label{sec:problem_formulation}

We consider a triplet input \((R_i, M_i, T_i)\), where \(R_i\) is a reference image, \(M_i\) is a modification text, and \(T_i\) is the target image.  
The goal of CIR is to learn a joint representation that reflects the transformation described by \(M_i\) when applied to \(R_i\) and closely aligns with \(T_i\). Specifically, we extract image and text features by applying \(f_\phi\) and \(f_\zeta\) to \(R_i\) and \(M_i\), respectively, where \(f_\phi\) is a pretrained image encoder and \(f_\zeta\) is a text encoder. These features are combined to form \(\boldsymbol{F_u}_{i}\), a fused representation derived from \(f_\phi(R_i)\) and \(f_\zeta(M_i)\). 
Our objective is to learn a fused representation that not only aligns globally with the target image embedding in the shared latent space, but also captures fine-grained semantic correspondences and remains robust under input variability. The overall training pipeline is illustrated in Fig.~\ref{fig:overview}.

\subsection{Enhanced Semantic Alignment Module}

ESAM is designed to capture both global semantic correlations between modalities and fine-grained semantic relationships between textual tokens and salient image regions. It adopts a late fusion paradigm as its foundation, which preserves modality-specific information and supports global alignment. However, conventional late fusion approaches often fail to capture detailed token-level correspondences. To overcome this, ESAM incorporates a cross-attention mechanism that explicitly models token-to-region alignments, complementing the high-level representation learning with fine-grained semantic consistency.

\subsubsection{Global Alignment via Contrastive Loss}
To realize global alignment, ESAM combines features extracted from the reference image \( f_\phi(R_i) \) and the textual modification \( f_\zeta(M_i) \). The fused representation is given by:
\begin{equation}
\boldsymbol{F_u}_{i} = \lambda_1 f_\phi(R_i) + (1 - \lambda_1) f_\zeta(M_i),
\end{equation}
where \( \lambda_1 \in [0, 1] \) controls the relative importance of each modality. The target image is encoded as \( f_\theta(T_i) = \boldsymbol{v}_i \). Then, \(\boldsymbol{F_u}_i\) is projected into a shared latent space through a multi-layer perceptron (\(\mathrm{MLP}\)).
 A contrastive loss based on cosine similarity is then used to align with the target:
\begin{equation}
\mathcal{L}_{\mathit{Late}} = -\frac{1}{\mathcal{B}} \sum_{i \in \mathcal{B}} 
\log \frac{\exp\bigl(\cos(\mathrm{MLP}(\boldsymbol{F_u}_i), \boldsymbol{v}_i)/\tau\bigr)}
          {\sum_{j \in \mathcal{B}} \exp\bigl(\cos(\mathrm{MLP}(\boldsymbol{F_u}_i), \boldsymbol{v}_j)/\tau\bigr)},
\end{equation} where \( \tau \) is a temperature parameter and $\mathcal{B}$ is the batch size.

\subsubsection{Fine-Grained Alignment via Cross-Attention}
Although $\mathcal{L}_{\mathit{Late}}$ supports high-level alignment, it fails to capture fine-grained interactions between modalities. To address this, ESAM utilizes a Q-Former mechanism that processes both (reference image + modification text) and (target image + modification text) in parallel to capture detailed token-to-region correspondences. Let $\boldsymbol{Q}_i$ be the query embedding derived from $f_\phi(R_i)$, and $\boldsymbol{K}_i$ the key embedding from $f_\zeta(M_i)$. A cross-attention map is computed as:
\begin{equation}
\boldsymbol{A}_i = 
\mathrm{softmax}\Bigl(\frac{\boldsymbol{Q}_i {\boldsymbol{K}_i}^T}{\sqrt{d}}\Bigr),
\end{equation}
where $d$ is the dimension of the query/key embeddings. This attention map $\boldsymbol{A}_i$ captures how each region (i.e., query) in $R_i$ aligns with the textual tokens. Similarly, for the target-image branch, we define the query embedding $\boldsymbol{Q}'_i$ from $f_\theta(T_i)$. Using this $\boldsymbol{Q}'_i$ and the same key embedding $\boldsymbol{K}_i$, we compute the corresponding attention map $\boldsymbol{A}'_i$ through the same cross-attention mechanism.
 After obtaining $\boldsymbol{A}_i$ and $\boldsymbol{A}'_i$, ESAM computes the following loss to align them:
\begin{equation}
\mathcal{L}_{\mathit{Attention}} = -\frac{1}{\mathcal{B}} \sum_{i \in \mathcal{B}} 
\log \frac{\exp\left(\cos\left({\boldsymbol{A}_i}^T, \boldsymbol{A}'_i\right) / \tau\right)}
         {\sum_{j \in \mathcal{B}} \exp\left(\cos\left({\boldsymbol{A}_j}^T, \boldsymbol{A}'_j\right) / \tau\right)}.
\end{equation}

In essence, we enforce similarity between the attention maps \(\boldsymbol{A}_i\) and \(\boldsymbol{A}'_i\), derived from \(R_i\) and \(T_i\) respectively, encouraging fine-grained consistency between them with respect to the same textual modification.

Through this two-branch attention design, ESAM captures fine-grained semantic correspondences that complement the global alignment enforced by $\mathcal{L}_{\mathit{Late}}$. Moreover, it ensures that changes specified by $T_i$ are consistently reflected when comparing reference and target images at the token-region level. The overall ESAM objective combines both loss terms:
\begin{equation}
\mathcal{L}_{\mathit{ESAM}} = \mathcal{L}_{\mathit{Late}} + \mathcal{L}_{\mathit{Attention}}.
\end{equation}


\begin{table*}[!t]
    \centering
    \caption{\textsc{Quantitative comparison across competing methods on FashionIQ. Avg. indicates the average results across all metrics for the three classes. The best results are marked in bold.}}
    \label{tab:comparison_results}
    {\fontsize{9.2pt}{11.8pt}\selectfont
    \begin{tabular}{>{\centering\arraybackslash}m{1.8cm}|c|cc|cc|cc|cc|c}
        \toprule
        \multirow{2}{*}{\textbf{Fusion}} & \multirow{2}{*}{\textbf{Method}} & \multicolumn{2}{c|}{\textbf{Dress}} & \multicolumn{2}{c|}{\textbf{Shirt}} & \multicolumn{2}{c|}{\textbf{Toptee}} & \multicolumn{2}{c|}{\textbf{Avg.}} & \multirow{2}{*}{\textbf{Avg.}} \\
        & & R@10 & R@50 & R@10 & R@50 & R@10 & R@50 & R@10 & R@50 & \\
        \midrule
        \multirow{4}{*}{Early}
        & SyncMask~\cite{song2024syncmask} & 33.76 & 61.23 & 35.82 & 62.12 & 44.82 & 72.06 & 38.13 & 65.14 & 51.64 \\
        & CASE~\cite{levy2024data} & 47.44 & 69.36 & 48.48 & 70.23 & 50.18 & 72.24 & 48.79 & 70.68 & 59.74 \\
        & SPRC~\cite{xu2024sentence} & 49.18 & 72.43 & 55.64 & 73.89 & \textbf{59.35} & 78.58 & 54.92 & 74.97 & 64.85 \\
        & PAIR~\cite{fu2025pair} & 46.78 & 70.93 & 52.60 & 73.80 & 58.91 & 78.81 & 52.76 & 74.51 & 63.64 \\
        \midrule
        \multirow{3}{*}{Late} 
        & CompoDiff~\cite{gu2024compodiff} & 40.65 & 57.14 & 36.87 & 57.39 & 43.93 & 61.17 & 40.48 & 58.57 & 49.53 \\
        & TG-CIR~\cite{wen2023target} & 45.22 & 69.66 & 52.60 & 72.52 & 56.14 & 77.10 & 51.32 & 73.09 & 58.05 \\
        & MEDIAN~\cite{huang2025median} & 46.90 & 70.30 & 52.65 & 73.96 & 57.62 & 78.63 & 52.39 & 74.30 & 63.34 \\
        \midrule
        Early + Late & Ours & \textbf{49.53} & \textbf{72.53} & \textbf{56.67} & \textbf{74.93} & 58.64 & \textbf{79.14} & \textbf{54.95} & \textbf{75.53} & \textbf{65.24} \\
        \bottomrule
    \end{tabular}
    }
\end{table*}

This formulation enables ESAM to effectively unify global alignment from late fusion with fine-grained semantic modeling via cross-attention, ensuring a robust and comprehensive multimodal representation.

\subsection{Adaptive Reconciliation Module}

Early fusion strategies enable direct integration of modalities at early stages, promoting flexible cross-modal interactions and unified representation learning. However, they often suffer from the loss of modality-specific nuances and reduced robustness under variable input conditions \cite{chen2022composed}. To leverage the strengths of early fusion while mitigating its limitations, we propose ARM, which incorporates two key mechanisms to address these issues. First, it introduces uncertainty-aware perturbations to improve robustness against noise and data variability.  
Second, it leverages the intermediate representations \(\boldsymbol{p}_i\) from ESAM to refine multimodal representations through alignment with target image embeddings.

\subsubsection{Handling Variability with Uncertainty-Aware Perturbations}
To address the sensitivity of early fusion to noise and data variability, ARM incorporates a robustness mechanism called \emph{Resilience Loss}. This mechanism combines deterministic embeddings \(\boldsymbol{v}_i\), which capture stable fused semantics, and perturbed embeddings \(\hat{\boldsymbol{v}}_i\), which employ an uncertainty-based weighting approach to preserve fine-grained semantic cues under diverse retrieval conditions. Concretely, we define a generic retrieval objective \(\mathcal{L}(\boldsymbol{v}^*_i)\) as follows:
\begin{equation}
\mathcal{L}\bigl(\boldsymbol{v}^*_i\bigr) = -\frac{1}{\mathcal{B}} \sum_{i \in \mathcal{B}} 
\log \frac{\exp\left({\boldsymbol{u}_i}^T \boldsymbol{v}^*_i / \tau\right)}
         {\sum_{j \in \mathcal{B}} \exp\left({\boldsymbol{u}_j}^T \boldsymbol{v}^*_j / \tau\right)},
\end{equation} where \( \boldsymbol{v}^*_i\ \in \{\boldsymbol{v}_i, \hat{\boldsymbol{v}}_i\} \) and  $\boldsymbol{u}_i$ is obtained by encoding the fused representation $\boldsymbol{U}_i$ from ESAM. The terms \(\mathcal{L}_{\text{Early}}\)  and \(\mathcal{L}_{\text{Uncertainty}}\) correspond to applying the loss \(\mathcal{L}\bigl(\boldsymbol{v}^*_i\bigr)\) with \(\boldsymbol{v}_i\) and \(\hat{\boldsymbol{v}}_i\), respectively. The perturbed embeddings \(\hat{\boldsymbol{v}}_i\) are computed as \(\hat{\boldsymbol{v}}_i = \alpha \boldsymbol{v}_i + \beta\), where \(\alpha\) and \(\beta\) are noisy vectors with the same shape as \(\boldsymbol{v}_i\), sampled from \(\mathcal{N}(1, \sigma_t)\) and \(\mathcal{N}(\mu_t, \sigma_t)\) respectively. Here, \(\mu_t\) represents the mean of the original target feature distribution, and \(\sigma_t\) is its standard deviation. Finally, the Resilience Loss \(\mathcal{L}_{\mathit{Res}}\) is defined as:
\begin{equation}
\label{eq:resilience_combined}
\mathcal{L}_{\mathit{Res}} 
      = \lambda_2\,\mathcal{L}_{\mathit{Early}}
        + \bigl(1 - \lambda_2\bigr)\,\mathcal{L}_{\mathit{Uncertainty}}.
\end{equation}

By jointly optimizing both deterministic and perturbed embeddings, this formulation enhances the model's ability to maintain consistency in fused representations while adapting to noise and variability.


\subsubsection{Bridging Text Prompts to Visual Targets}

To ensure that the retrieval process remains faithful to the intended textual modification, ARM introduces an additional objective called \emph{Prompt-to-Image Loss}. This loss leverages the intermediate representation \(\boldsymbol{p}_i\), which encodes the semantic intent of the modification, and aligns it with the final target image embedding \(\boldsymbol{v}_i\). The prompt embedding is projected into the shared latent space using the text encoder as \(\boldsymbol{u}'_i = f_\zeta(\boldsymbol{p}_i)\), and optimized using the following contrastive formulation:
\begin{equation}
\mathcal{L}_{\mathit{PI}} = -\frac{1}{\mathcal{B}} \sum_{i \in \mathcal{B}} 
\log \frac{\exp\left({\boldsymbol{u}'_i}^T \boldsymbol{v}_i / \tau\right)}
         {\sum_{j \in \mathcal{B}} \exp\left({\boldsymbol{u}'_j}^T \boldsymbol{v}_j / \tau\right)}.
\end{equation}

This objective contributes to more accurate and semantically faithful alignment by improving the representational consistency of the query, through semantic alignment between the text prompt and its visual target.

ARM integrates Resilience loss and Prompt-to-Image loss to refine multimodal representations while balancing robustness and alignment precision:
\begin{equation}
\mathcal{L}_{\mathit{ARM}} = \mathcal{L}_{\mathit{Res}} + \mathcal{L}_{\mathit{PI}}.
\end{equation}

This approach ensures that ARM generalizes effectively across diverse retrieval scenarios, handling noise and variability while maintaining precise alignment between modalities. The total learning objective of the FAR-Net combines the contributions of ESAM and ARM:
\begin{equation}
\mathcal{L}_{\mathit{Total}} = \mathcal{L}_{\mathit{ESAM}} + \mathcal{L}_{\mathit{ARM}}.
\end{equation}

This comprehensive strategy combines the high-level correlations and fine-grained semantic relationships captured by ESAM with the robustness and adaptability provided by ARM, enabling both high retrieval accuracy and robust generalization under challenging conditions.

\begin{table*}[!t]
    \centering
    \caption{\textsc{Quantitative comparison on CIRR. The best results are in bold. Avg. indicates }$(\text{R@5} + \text{R}_{\textit{subset}}\text{@1})/2$.}
    \label{tab:cirr_results}
    {\fontsize{9.2pt}{11.8pt}\selectfont
    \begin{tabular}{>{\centering\arraybackslash}m{1.8cm}|c|cccc|ccc|c}
        \toprule
        \multirow{2}{*}{\textbf{Fusion}} & \multirow{2}{*}{\textbf{Method}} 
        & \multicolumn{4}{c|}{\textbf{R@K}} 
        & \multicolumn{3}{c|}{\textbf{R\textsubscript{\textit{subset}}@K}} 
        & \multirow{2}{*}{\textbf{Avg.}} \\
        \cmidrule(lr){3-6} \cmidrule(lr){7-9}
        & & R@1 & R@5 & R@10 & R@50 & R@1 & R@2 & R@3 & \\
        \midrule
        \multirow{3}{*}{Early} 
        & CASE~\cite{levy2024data}     & 48.00 & 79.11 & 87.25 & 97.57 & 75.88 & 90.58 & 96.00 & 77.50 \\
        & SPRC~\cite{xu2024sentence}   & 51.96 & 82.12 & 89.74 & 97.69 & 80.65 & 92.31 & 96.60 & 81.39 \\
        & PAIR~\cite{fu2025pair}       & 46.36 & 78.43 & 87.86 & \textbf{97.90} & 74.63 & 89.64 & 95.61 & 76.53 \\
        \midrule
        \multirow{4}{*}{Late}
        & CompoDiff~\cite{gu2024compodiff} & 22.35 & 54.36 & 73.41 & 91.77 & 35.84 & 56.11 & 76.60 & 45.10 \\
        & TG-CIR~\cite{wen2023target}     & 45.25 & 78.29 & 87.16 & 97.30 & 72.84 & 89.25 & 95.13 & 75.59 \\
        & SSN~\cite{yang2024decomposing}  & 43.91 & 77.25 & 86.48 & 97.45 & 71.76 & 88.63 & 95.54 & 74.51 \\
        & MEDIAN~\cite{huang2025median}   & 45.66 & 78.72 & 87.88 & 97.89 & 75.52 & 89.45 & 95.57 & 77.12 \\
        \midrule
        Early + Late & Ours & \textbf{54.39} & \textbf{83.06} & \textbf{90.60} & 97.83 & \textbf{80.77} & \textbf{92.41} & \textbf{96.68} & \textbf{81.92} \\
        \bottomrule
    \end{tabular}
    }
\end{table*}

\section{EXPERIMENTS}\label{sec:sota_comparison}

\subsection{Datasets and Evaluation Protocol}
We evaluate our model on two widely used datasets, FashionIQ and CIRR. FashionIQ comprises three fashion categories (Dress, Shirt, and Toptee) with 46k training images and 15k images for validation and testing. It includes 18k training queries and 12k queries each for validation and testing, where each query consists of two captions describing modifications from the reference to the target image. CIRR consists of over 36k open-domain images annotated with human-generated textual modifications. Following \cite{luo2022clip4clip}, we split CIRR into 80\% training, 10\% validation, and 10\% testing. For evaluation, we adopt Recall@K (R@K) as in \cite{delmas2022artemis}. Additionally, for CIRR we report Recall@K on the visually similar subset (\textnormal{R\textsubscript{\textit{subset}}@K}) to measure fine-grained vision-language understanding.

\begin{table}[t]
    \centering
    \caption{\textsc{Ablation studies of FAR-Net with various settings on FashionIQ and CIRR.}}
    \label{tab:ablation}
    {\fontsize{9.2pt}{11.8pt}\selectfont  
    \begin{tabular}{c|c|c}
        \toprule
        \multirow{2}{*}{\textbf{Method}} & \textbf{FashionIQ} & \textbf{CIRR} \\
        & \textbf{Avg.} & \textbf{Avg.}  \\
        \midrule
        w/o $\mathcal{L}_{\mathit{Late}}$        & 46.28 & 63.64 \\
        w/o $\mathcal{L}_{\mathit{Attention}}$   & 57.24 & 67.11 \\
        w/o $\mathcal{L}_{\mathit{PI}}$          & 59.45 & 74.13 \\
        w/o $\mathcal{L}_{\mathit{Res}}$         & 62.06 & 78.45 \\
        \midrule
        ESAM only                              & 62.44 & 79.28 \\
        ARM only                             & 40.18 & 66.33 \\
        \midrule
        \textbf{FAR-Net}                         & \textbf{65.24} & \textbf{81.92} \\
        \bottomrule
    \end{tabular}
    }
\end{table}

\subsection{Implementation Details} 
We adopt a pre-trained ViT-L \cite{dosovitskiy2020image} as the image encoder\cite{ lee1999integrated, roh2007accurate}. The text encoder is sourced from BLIP-2, based on BERT \cite{devlin2019bert}. We use a batch size of 32 and resize all images to \(224 \times 224\) with a padding ratio of 1.25 for uniformity. Training is conducted for 50 epochs with AdamW \cite{loshchilov2017decoupled}, using a weight decay of 0.05. The model is trained on an NVIDIA RTX A40 GPU (48GB).

\subsection{Comparison with State-of-the-Art Methods}
As shown in Table~\ref{tab:comparison_results}, on the fashion-oriented FashionIQ, our method achieves higher averages in R@10 and R@50 than previous state-of-the-art approaches. Moreover, it achieves the highest overall performance on the real-world CIRR, with results summarized in Table~\ref{tab:cirr_results}. In particular, our model surpasses the best baseline by 2.43\% in R@1, indicating the most accurate retrieval capability. These improvements on FashionIQ and CIRR highlight our model’s robust retrieval performance across diverse domains.

\subsection{Extended Experiments and Analysis} 

\subsubsection{Ablation Study}
From our ablation results, removing $\mathcal{L}_{\mathrm{Late}}$ decreases the average retrieval performance by 18.96\% on FashionIQ and by 18.28\% on CIRR, highlighting the necessity of global alignment.
\begin{figure}[t]
    \centering
    \includegraphics[width=0.48\textwidth]{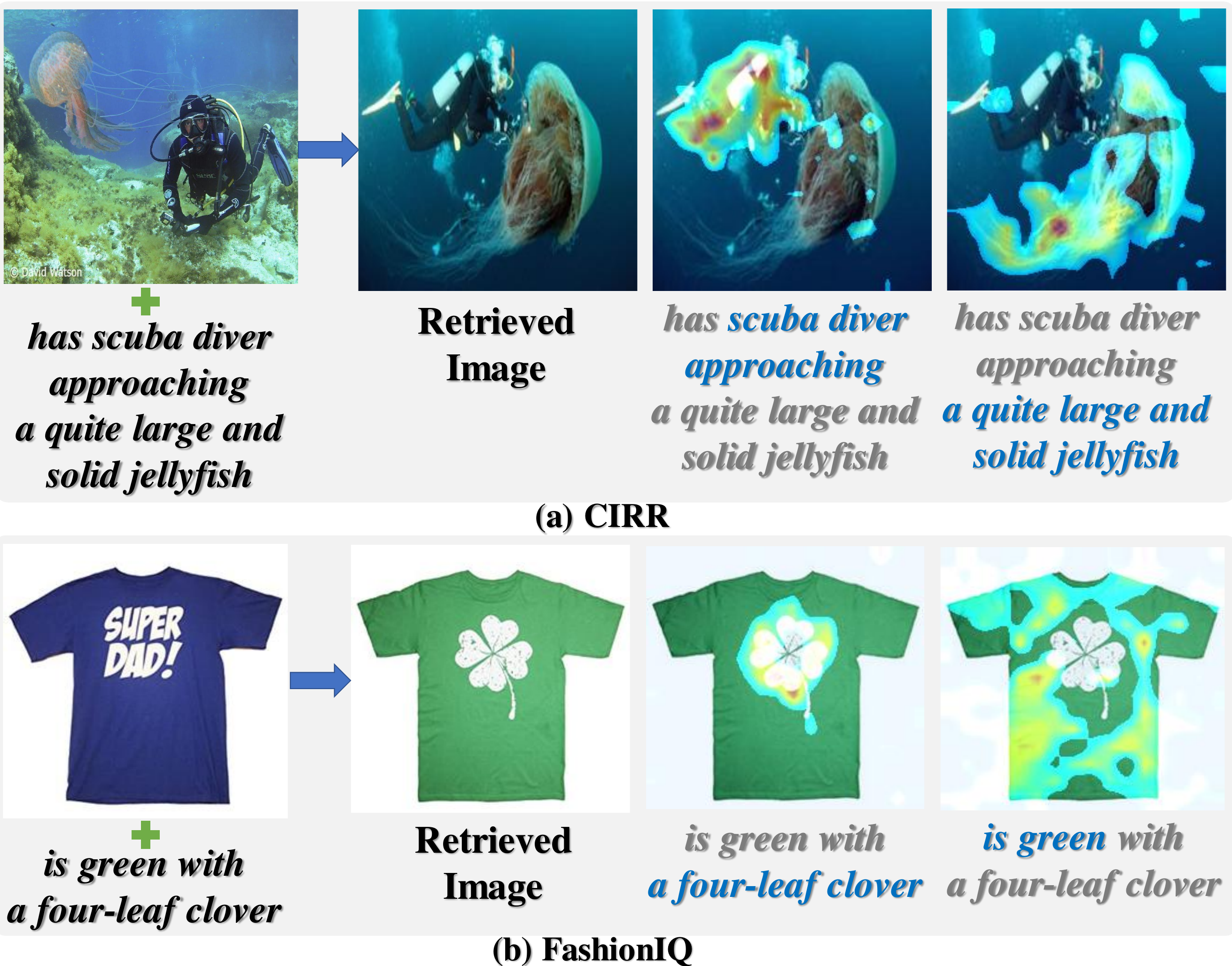}
    \caption{Attention visualization for semantic alignment on (a) CIRR and (b) FashionIQ.}
    \label{fig:attention}
\end{figure}
Likewise, discarding $\mathcal{L}_{\mathrm{Attention}}$ lowers the FashionIQ performance by 8.00\%, showing the impact of fine-grained token-to-region alignment. Further, removing $\mathcal{L}_{\mathrm{PI}}$ or $\mathcal{L}_{\mathrm{Res}}$ results in a drop of 5.79\% and 3.18\%, respectively, indicating their contributions to semantic consistency and robustness. In contrast, combining all modules (ESAM + ARM) achieves the best overall performance, confirming the synergy between enhanced cross-attention and uncertainty modeling for composed image retrieval.

\subsubsection{Visualization of Semantic Alignment}
To qualitatively assess FAR-Net's fine-grained semantic alignment capability, we visualize cross-attention map highlights as shown in Fig.~\ref{fig:attention}. In both CIRR and FashionIQ examples, FAR-Net precisely identifies regions corresponding to specific textual modifications. Specifically, in the CIRR example in Fig.~\ref{fig:attention}(a), the highlighted regions distinctly cover the relevant image region, closely matching the provided modification text. Similarly, in the FashionIQ example in Fig.~\ref{fig:attention}(b), the regions corresponding to the described attributes, such as ``green color'' and ``four-leaf clover'', are prominently emphasized. These visualizations demonstrate FAR-Net's ability to accurately align nuanced textual instructions with relevant visual regions, reinforcing the effectiveness of our proposed semantic alignment approach.

\section{CONCLUSION}

In this paper, we propose FAR-Net, a text-image fusion framework evaluated on the fashion-centric FashionIQ and the more diverse CIRR benchmark. Our model achieves strong retrieval accuracy even under CIRR's broader contextual relationships, indicating its generalizability beyond narrowly defined domains. On FashionIQ, it excels at capturing nuanced visual and textual cues, highlighting its effectiveness in fine-grained fashion scenarios. Our contributions are threefold. First, FAR-Net adopts a multi-stage architecture that integrates the complementary strengths of early and late fusion strategies through semantic alignment and adaptive reconciliation. Second, it leverages cross-attention and uncertainty modeling to enhance fine-grained semantic alignment and robustness. Lastly, comprehensive experiments across FashionIQ and CIRR confirm FAR-Net’s ability to handle diverse and complex multimodal retrieval tasks. 

Although the current framework is trained in a supervised setting, a natural extension is to explore zero-shot adaptation or domain transfer to reduce reliance on large-scale annotations. Evaluating FAR-Net under zero-shot or cross-domain scenarios would further validate its robustness and broad applicability.

\bibliographystyle{IEEEtran}
\bibliography{references}

\end{document}